\documentclass{article} 
\usepackage[preprint]{colm2025_conference}

\usepackage{microtype}
\usepackage{hyperref}
\usepackage{url}
\usepackage{booktabs}

\usepackage{lineno}

\definecolor{darkblue}{rgb}{0, 0, 0.5}
\hypersetup{colorlinks=true, citecolor=darkblue, linkcolor=darkblue, urlcolor=darkblue}

\usepackage{siunitx}
\usepackage{graphicx}
\usepackage{enumitem}       
\usepackage{amsmath}
\usepackage{multirow}
\usepackage{booktabs}
\usepackage{array}
\usepackage{subfigure}
\usepackage{amssymb,mathtools}
\usepackage{verbatim}
\usepackage{bm}     
\usepackage{enumitem}       
\usepackage{arydshln} 
\usepackage{subcaption}
\usepackage{caption} 
\usepackage[normalem]{ulem} 
\title{TransMamba: A Sequence-Level Hybrid Transformer-Mamba Language Model}

\author{\textbf{Yixing Li\textsuperscript{2}$\dagger$, Ruobing Xie\textsuperscript{1}$\ast$, Zhen Yang\textsuperscript{1}, Xingwu Sun\textsuperscript{1,3}, Shuaipeng Li\textsuperscript{1}, Weidong Han\textsuperscript{1},}  \\
\textbf{Zhanhui Kang\textsuperscript{1}, Yu Cheng\textsuperscript{2}$\ast$, Chengzhong Xu\textsuperscript{3}, Di Wang\textsuperscript{1}, Jie Jiang\textsuperscript{1}}
\\
 \textsuperscript{1}Tencent Hunyuan
 \\
 \textsuperscript{2}The Chinese University of Hong Kong 
 \\
 \textsuperscript{3}University of Macau
 \\
\texttt{li.yixing@outlook.com}~~~
\texttt{xrbsnowing@163.com}~~~
\\
\texttt{\{andreasyang, sammsun, jonnyhan\}@tencent.com}~~~
\texttt{chengyu@cse.cuhk.edu.hk}
}

\begin{document}

\ifcolmsubmission
\linenumbers
\fi

\maketitle

\begin{abstract}


Transformers are the cornerstone of modern large language models, but their quadratic computational complexity limits efficiency in long-sequence processing. Recent advancements in Mamba, a state space model (SSM) with linear complexity, offer promising efficiency gains but suffer from unstable contextual learning and multitask generalization. Some works conduct layer-level hybrid structures that combine Transformer and Mamba layers, aiming to make full use of both advantages. This paper proposes TransMamba, a novel sequence-level hybrid framework that unifies Transformer and Mamba through shared parameter matrices (QKV and CBx), and thus could dynamically switch between attention and SSM mechanisms at different token lengths and layers. We design the Memory Converter to bridge Transformer and Mamba by converting attention outputs into SSM-compatible states, ensuring seamless information flow at TransPoints where the transformation happens. The TransPoint scheduling is also thoroughly explored for balancing effectiveness and efficiency. We conducted extensive experiments demonstrating that TransMamba achieves superior training efficiency and performance compared to single and hybrid baselines, and validated the deeper consistency between Transformer and Mamba paradigms at sequence level, offering a scalable solution for next-generation language modeling. Code and data are available at https://github.com/Yixing-Li/TransMamba.

\end{abstract}

\section{Introduction}
\let\thefootnote\relax\footnotetext{$\ast$ Corresponding author.}
\let\thefootnote\relax\footnotetext{$\dagger$ Work conducted during internship at Tencent.}

Transformers \citep{vaswani2017attention, achiam2023gpt, touvron2023llama} are the foundation and mainstream model of modern deep learning \citep{zhao2023survey}, showing dominating power in language modeling. 
Recently, Mamba has emerged \citep{mamba} and been verified in various fields. Compared with Transformer, Mamba has linear computational complexity, high efficiency in processing long sequences, and lower training and inference costs \citep{qu2024survey}. Nevertheless, its contextual learning and multi-task generalization capabilities are unstable \citep{waleffe2024empirical}. Transformer and Mamba have their own strengths and complement each other.

However, Transformer and Mamba have their own flaws that cannot be addressed by naive layer-shared hybrid structures \citep{yuan2024remamba, yang2024efficient}. For example, Transformer has faster training for short contexts while Mamba has better efficiency in longer contexts (see Table \ref{tab:efficiency_attn_SSM}).
Moreover, the naive static hybrid model has structural restrictions such as the order of Mamba and Transformer, mandatory ratios, etc \citep{lieber2024jamba, Mamba2}. The performance of the Hybrid model will deteriorate if these specific rules are not met, which greatly limits the exploration and breakthrough of the model.

\begin{figure}[!t]
    \centering 
  \includegraphics[width=0.82\linewidth]{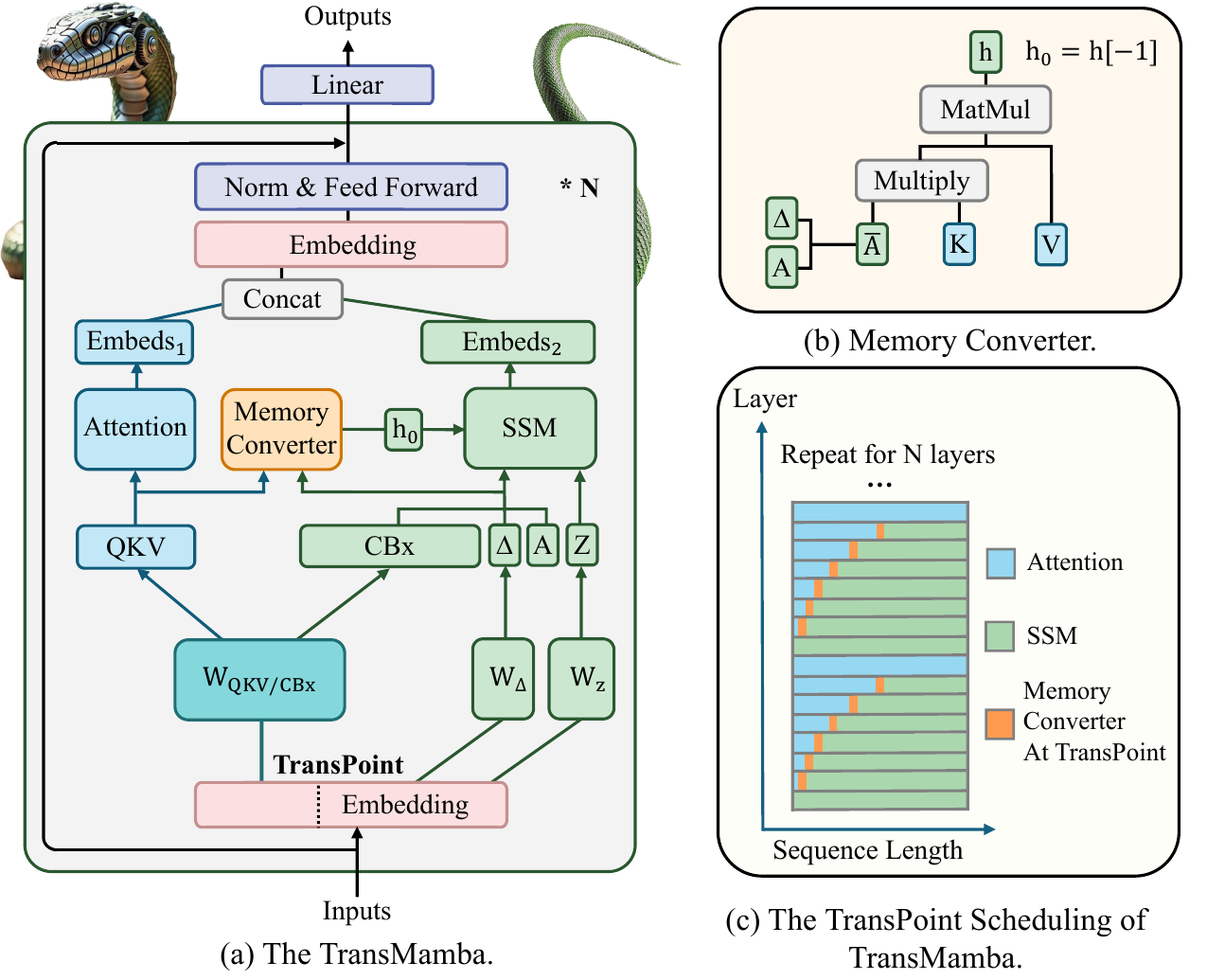} \hfill
  \caption {(a) Structure of TransMamba. Attention and SSM have shared parameters $\mathbf{W_{QKV}}$ and $\mathbf{W_{CBx}}$. Tokens are either processed via the green path (SSM mode) or the blue path (Attention mode). (b) Memory Converter. (c) The TransPoint Scheduling of TransMamba.}
  \label{fig:TransMamba1}
\end{figure}

Recently, Mamba2 \citep{Mamba2} further enhances the performance of Mamba series, which reveals the surprising consistency of the attention of Transformer and the State Space Model (SSM) of Mamba. 
Furthermore, \citep{distill_mamba} performed distillation between Mamba and Transformer, distilling the QKV parameters of Attention to obtain CBx of SSM, verifying that the parameters can be interactively transferred as shown in Table \ref{tab:consistency}.
These motivate us that we can bravely utilize a set of shared parameters of QKV and CBx to build a joint Transformer-Mamba framework, which could \emph{flexibly decide which structure is suitable for the current training/inference in different layers/token lengths}, taking advantage of both structures to balance effectiveness and efficiency while ensuring structural flexibility.
Intuitively, to obtain the efficiency advantages of both structures, we can make the model adopt the Transformer mechanism for training on relatively short contexts and the SSM mechanism on long contexts.
As shown in Figure \ref{fig:TransMamba1}, such prototype framework has only one set of parameters to flexibly switch between Transformer and Mamba for LM. In the first N tokens of the sequence, the parameter matrix is calculated using the attention mechanism. At a specific node in the sequence (which we call \textbf{\emph{TransPoints}} from Transformer mode to Mamba2 mode), the parameter matrix is converted to the SSM mechanism for subsequent sequence generation, so as to achieve better training efficiency with better performance in sequences of different lengths.



\begin{figure}[t]
    \centering
    \begin{minipage}[t]{0.63\textwidth}
        \centering
        \includegraphics[width=\textwidth]{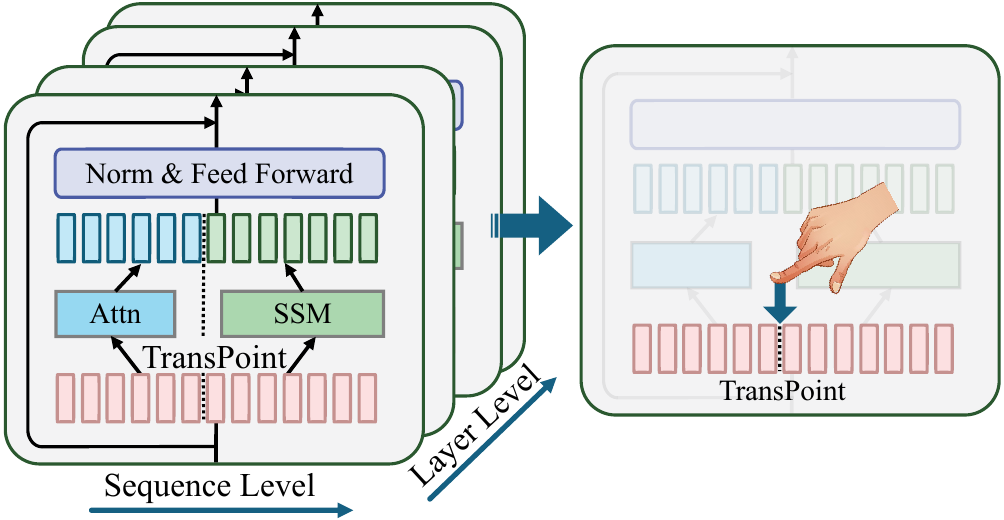}
        \caption{TransMamba has shared parameters to flexibly switch between Attention and SSM, and TransPoints decide which parts of token sequence use Attention or SSM.}
        \label{method:fig1-1}
    \end{minipage}
    \hfill
    \begin{minipage}[t]{0.35\textwidth}
        \centering
        \includegraphics[width=\textwidth]{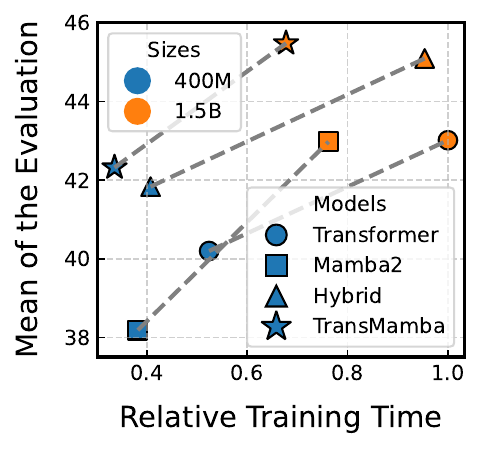}
        \caption{TransMamba generally shows better efficiency and performance with different sizes.}
        \label{method:fig1-2}
    \end{minipage} \label{fig:TransMamba generally shows better}

\end{figure}


The implementation of this flexible token-level Transformer-Mamba transformation is non-trivial and has the following challenges: (1) In the TransPoints between, the latter structure (Mamba) should well capture the information of the previous tokens learned by the former structure (Transformer) via an appropriate method that the latter structure could understand. How to losslessly transfer the knowledge learned by the previous Transformer to the latter SSM modeling part is essential. (2) We could flexibly decide when (e.g., at what sequence length) to transfer from Transformer to Mamba at different layers in such framework. Jointly considering effectiveness and efficiency, the selection of a reasonable set of TransPoints requires careful explorations under insightful principles. (3) The structures of this framework varies at different sequence lengths (e.g., pure Transformer/Mamba2 or certain Hybrid structures), in which case the model performance should be concerned.


To address these problems, we propose a novel \textbf{\emph{TransMamba}} framework that utilizes the same set of shared parameters to flexibly switch between attention and SSM mechanisms in token generation at different sequence lengths and layers, combining the advantages in effectiveness and efficiency of Transformer and Mamba.
Specifically, we design a sophisticated \textbf{\emph{Memory Converter}} to convert the intermediate results of the attention part into the state required by the SSM mechanism, ensuring the consistency of the information around the TransPoint with tokens being processed, and no loss will be incurred when converting between attention and SSM.
Moreover, we have conducted comprehensive research on the \textbf{\emph{TransPoint schedule}}, exploring the overall optimal TransPoint setting and insights in different layers and sequence lengths. In this case, our TransMamba framework could be viewed as a flexible dynamic combination of hybrid Transformer/Mamba layers varies in different token lengths.
Our contributions are summarized as follows:
\begin{itemize}[leftmargin=0.36cm]
    \item We propose a novel TransMamba framework, which verifies the consistency of Transformer and Mamba in a deeper degree, starting from the one shared set of parameters while outputting tokens via two different mechanisms.
    \item We design the Memory Converter that conforms to the theoretical solution to ensure the consistency of information in TransMamba during the conversion process, and explore the optimal TransPoint schedule at different layers and token lengths.
    \item We conduct extensive experiments to verify the performance and efficiency advantages of TransMamba on both effectiveness and efficiency. In conclusion, TransMamba could be a promising structure for LM.
\end{itemize}



\section{Method}

\subsection{Preliminary}

\begin{table}[t]
    \centering
    \small
  \begin{minipage}{0.45\textwidth}
    \centering
    \renewcommand\arraystretch{1.23}
\begin{tabular}{ll}

\hline  Attention & SSM \\
\hline $\mathbf{Q} = \delta(\mathbf{H} \mathcal{W}_{\mathbf{Q}})$  & $\mathbf{C} = \delta(\mathbf{H} \mathcal{W}_{\mathbf{C}})$  \\
 $\mathbf{K} = \delta(\mathbf{H} \mathcal{W}_{\mathbf{K}})$ & $\mathbf{B} = \delta(\mathbf{H} \mathcal{W}_{\mathbf{B}})$   \\
  $\mathbf{V} = \delta(\mathbf{H} \mathcal{W}_{\mathbf{V}})$& $\mathbf{X} = \delta(\mathbf{H} \mathcal{W}_{x}) \circ \Delta $  \\
  $\mathbf{y} = (\mathbf{L} \circ \mathbf{Q} \mathbf{K}^{T})\mathbf{V}$ & $\mathbf{y} = (\mathbf{A}^{\times} \circ \mathbf{C} \mathbf{B}^{T})\mathbf{X}$  \\
\hline 
\end{tabular}
    \caption{Compare the matrix form of SSM and Attention. The core mechanisms of Attention and SSM show consistency in dual form, which is the mathematical basis that enables us to unify Transformer and Mamba. }\label{tab:consistency} 
  \end{minipage}%
  \hfill
  \begin{minipage}{0.53\textwidth}
    \centering
    \renewcommand\arraystretch{1.21}
  {
    \begin{tabular}{lc}
    \hline Model & Training FLOPs   \\ 
    & / Layer   \\
    \hline Transformer & $O(\mathrm{T}^2 \mathrm{N})$  \\
    Mamba & $O(\mathrm{TN}^2)$  \\
    TransMamba & $O(\mathrm{P}^2\mathrm{N}+(\mathrm{T-P})\mathrm{N}^2)$  \\
    \hline
    \end{tabular}
  }
    \caption{Compare the training FLOPs of Transformer, Mamba and optimal TransMamba. The FLOPs of TransMamba is a quadratic function of the TransPoint, and its specific value is related to the speed optimization coefficients of Transformer and Mamba respectively.}  \label{tab:efficiency_attn_SSM} 
  \end{minipage}
\end{table}

\subsubsection{Basic Notions and Consistency of Attention and SSM}\label{sec:method:attn_SSM_same}

We use the classic notation from the Transformer and Mamba papers.  $\mathbf{QKV}$ denotes the key parameters (query, key, value) of Attention, and  $\mathbf{L}$ denotes the additional mask matrix.  $\mathbf{CBx}$ represents the key parameters in the SSM,  $\Delta$ is used to control the discrete step size in SSM, and  $\mathbf{A}$ is used to describe the global dependencies of the hidden state, which is similar to the mask matrix in Attention \citep{mamba}. In order to satisfy the classic symbolic representation and clear expression, we use $\mathbf{H}$ to denote the input embeddings for attention and SSM. The corresponding calculations are shown in Table \ref{tab:consistency}.

Mamba2 \citep{Mamba2} compares the underlying mechanisms of Transformer and Mamba, and introduces the dual form of SSM to illustrate the consistency between the two.
In Table \ref{tab:consistency}, we can find that the core mechanisms of Transformer and Mamba  (attention and SSM) are completely symmetrical.
\citet{distill_mamba} aligned the QKV of the transformer weights with CBx of Mamba and performed distillation, achieving improved results on chat and long-text benchmarks. This once again shows that the core weights of Transformer and Mamba are transferable and unified.
The above theories and research inspired us to build a bolder framework of TransMamba with a unified architecture of Transformer and Mamba.


\subsubsection{Efficiency of Attention and SSM with different token lengths}\label{sec:effiency_of_attn_SSM}




As shown in Table \ref{tab:efficiency_attn_SSM}, recent work \citep{Mamba2} theoretically summarizes the FLOPs of Attention and SSM. $\mathrm{T}$ denotes the sequence length, $\mathrm{N}$ denotes the state dimension and $\mathrm{P}$ denotes the TransPoint value. When T is greater than N, Transformer has an advantage in efficiency on shorter sequences, while Mamba is efficient at training on long sequences due to its linear complexity of T. 
This advantage of Mamba's training efficiency on longer contexts is present with most of the commonly-used model sizes, which also forms the motivation of our TransMamba that attempts to unleash the maximum potential of the flexible hybrid Transformer-Mamba structure in terms of effectiveness and efficiency.

\subsection{Overall Framework of TransMamba}

\noindent
\textbf{Main architecture.}
As shown in Figure \ref{fig:TransMamba1} (a), TransMamba is a layer-stacked Decoder-only autoregressive model. Each layer of TransMamba contains all the parameters of Mamba, including the parameters required to calculate C, B, x, A and $\Delta$. Based on the aforementioned consistency between Transformer and Mamba, we boldly let QKV and CBx share the same parameters (i.e., Q$\leftrightarrow$C, K$\leftrightarrow$B, V$\leftrightarrow$x). In other words, our model \textbf{\emph{has the ability to switch between Transformer and Mamba structures, but with only one set of parameters}}.

In addition, TransMamba contains the crucial Memory Converter used for lossless information conversion when model parameters are switched from QKV to CBx (in Section \ref{sec:method:memory_converter}), armed with our TransPoint schedule that decides whether we should use Attention mode or SSM mode at a certain layer or token length (in Section \ref{TransPoint Scheduling}). 
To ensure better training efficiency, we only set a single TransPoint (i.e., the token position where the switch from Attention to SSM or vice versa happens) for each layer. The sequence before the TransPoint is calculated using Attention, and the rest is calculated through SSM. Complex structures with multiple TransPoints may have more magical properties and effects, which can be provided for future research. At different token lengths, TransMamba could be flexibly regarded as different structures (e.g., pure Transformer, Mamba, or Hybrid Transformer-Mamba).


\noindent
\textbf{Formalized calculation process.}
We denote the hidden state of the input tokens as $\mathbf{h}$. The remaining critical mathematical symbols are the same as given in Section \ref{sec:method:attn_SSM_same}.
TransMamba calculates intermediate results through linear project and convolution modules. Since the parts of the input token sequence that are shorter and longer than the TransPoint will be calculated through different mechanisms, for the sake of clarity, we use different symbols to represent the two parts:
(a) For the relatively former part of the input before TransPoint:
\begin{equation}
    \begin{aligned}
    &\mathbf{h_s} = \mathbf{h[:TransPoint]}, \quad
&\mathbf{Q} = \delta(\mathbf{h_s} \mathcal{W}_{\mathbf{C}}), \\
&\mathbf{K} = \delta(\mathbf{h_s} \mathcal{W}_{\mathbf{B}}), \quad
&\mathbf{V} = \delta(\mathbf{h_s} \mathcal{W}_{\mathbf{x}}). 
\end{aligned}
\end{equation}
The output $\mathbf{y_s}$ will be calculated through the attention mechanism before TransPoint:
\begin{equation}
    \begin{aligned}
\mathbf{y_s} &= \text{softmax}(\mathbf{Q} \mathbf{K}^{T})\cdot \mathbf{V}.
\end{aligned}
\end{equation}
(b) For the relatively latter part of the input after TransPoint:
\begin{equation}
    \begin{aligned}
    &\mathbf{h_l} = \mathbf{h[TransPoint:]}, \quad
&\Delta = \sigma(\mathbf{h_l} \mathcal{W}_{\Delta} + b_{\Delta}), \\
&\overline{\mathbf{A}} = e^{- \Delta e^{\log{\mathcal{W}_{\mathbf{A}}}}}, \quad
&\mathbf{C} = \delta(\mathbf{h_l} \mathcal{W}_{\mathbf{C}}),\\
&\mathbf{B} = \delta(\mathbf{h_l} \mathcal{W}_{\mathbf{B}}), \quad
&\mathbf{x} = \delta(\mathbf{h_l} \mathcal{W}_{\mathbf{x}}).
\end{aligned}
\end{equation}
TransMamba utilizes the SSM mechanism to generate outputs $\mathbf{y_l}$ after TransPoint. 
The initial state $h_{0}$ will be obtained through the Memory Converter:
\begin{equation}
    \begin{aligned}
    &h_{0} = \text{Memory Converter}(\mathbf{K}, \mathbf{V}), \quad
    &y_{k} = \mathbf{C}_{k} h_{k}, \\
    &h_{k} = \overline{\mathbf{A}_{k-1}} h_{k-1} + \mathbf{B}_{k}, \Delta_{k} x_{k}, \quad
    &\mathbf{y_l} = [y_{0}, \cdots, y_{k}].\\
\end{aligned}
\end{equation}
or in the matrix form:
\begin{equation}
    \mathbf{y_l} = (\mathbf{A}^{\times} \circ \mathbf{C} \mathbf{B}^{T}) (\Delta \circ \mathbf{x}).
\end{equation}
The final output of our TransMamba can be expressed as the combination of $\mathbf{y_s},\mathbf{y_l}$:
\begin{equation}
    \begin{aligned}
\mathbf{y} & = [\mathbf{y_s}, \mathbf{y_l}]. 
\end{aligned}
\end{equation}


\textbf{Feasibility.}
The feasibility of our design comes from two key points: (1) Due to the consistency of the attention and SSM mechanisms described in Section \ref{sec:method:attn_SSM_same}, the output of TransMamba can be calculated either through the attention or SSM mechanism flexibly. 
(2) Due to the power of our memory converter, TransMamba does not lose any information when converting from attention to SSM as the token length increases across the TransPoint. The sequence state required by SSM can be perfectly preserved by the $\mathbf{K}$ and $\mathbf{V}$ of attention.


\subsection{Lossless Memory Converter}\label{sec:method:memory_converter}

The memory converter is aimed to losslessly convert $\mathbf{K}$ and $\mathbf{V}$ calculated before TransPoint into the hidden state $\mathbf{h}$ required for the Mamba mode after TransPoint.
First, we expand the mathematical form of SSM in detail:
\begin{equation}
\begin{aligned}
&\Delta_{k} = \sigma(x_{k} \mathcal{W}_{\Delta} + b_{\Delta}), \quad
&\overline{\mathbf{A}_{k}} = e^{- \Delta_{k} e^{\log{\mathcal{W}_{\mathbf{A}}}}}, \\
&\mathbf{B}_{k} = \delta(x_{k} \mathcal{W}_{\mathbf{B}}), \quad
&\mathbf{C}_{k} = \delta(x_{k} \mathcal{W}_{\mathbf{C}}), \\
&h_{0} = \mathbf{B}_{0} \Delta_{0} x_{0}, \quad
&h_{k} = \overline{\mathbf{A}_{k-1}} h_{k-1} + \mathbf{B}_{k} \Delta_{k} x_{k}.
\end{aligned}
\end{equation}
Abbreviate $h$ to matrix form as:
\begin{equation}
\begin{aligned}
h =(\mathbf{A}^\times \circ \mathbf{B}^T)(\Delta \circ \mathbf{x}) = (\mathbf{A}^\times \circ \mathbf{B}^T) \mathbf{X},
\end{aligned}
\end{equation}\label{eq:used_A}
where $\mathbf{A}^{\times}$ is the lower triangular matrix obtained by arranging the elements of $\overline{\mathbf{A}}$, and details are shown in Appendix \ref{app:method_details_MemoryConverter_1}. Based on the consistency of the mathematical structure of attention and SSM shown in Section \ref{sec:method:attn_SSM_same}, we can calculate the estimated hidden state from the intermediate results K, V of attention as follows:
\begin{equation}
    \begin{aligned}
    h_{s} &=(\mathbf{A}^\times \circ \mathbf{K}^T) \mathbf{V}.
\end{aligned}
\end{equation}
The initial state of the TransPoint can be obtained as $h_0 = h_s[-1]$.
Therefore, TransMamba can transform losslessly from attention to SSM during sequence generation. It should be noted that our Memory converter does not require additional parameters, but is a theoretical solution calculated from existing results.

\subsection{Flexible TransPoint Scheduling}\label{TransPoint Scheduling}

TransPoint represents the token position of the segmentation of the sequence where the Transformer$\rightarrow$Mamba mode switch happens via the above Memory converter for each layer.
The position of TransPoint in the sequence can control the ratio of attention and SSM in this layer of TransMamba. For example, when TransPoint is set to the midpoint of the sequence, this layer is a 1:1 combination of Transformer and Mamba in the sequence level; if it is set to the beginning of the sequence, this layer is equal to Mamba. The TransPoint schedule decides the functional (hybrid) structure of TransMamba at different token lengths. 



\subsubsection{Principles of TransPoint Scheduling}


The TransPoint Scheduling aims to to maximize the respective advantages of Attention and SSM in short and long context training to optimize the overall efficiency and performance.
In summary, TransPoint scheduling should meet the following requirements:
\begin{itemize}[leftmargin=0.36cm]
    \item TransPoint has a great impact on training time. The distribution of TransPoints can be closer to the optimal position in Table \ref{tab:efficiency_attn_SSM} for better training efficiency;
    \item TransPoints at different layers cannot be too concentrated at one position. Under the premise of the first point, it needs to be distributed over the entire length of the sequence to prevent possible degradation brought by the mutations of simultaneous Transformer-to-Mamba transformations for better effectiveness;
    \item Due to the asynchronous transformations, our TransMamba could be viewed as different hybrid Transformer-Mamba structures at different token lengths. Therefore, we should take fully advantages of the superior hybrid Transformer-Mamba structures' insights to further enhance the effectiveness.
\end{itemize}

\subsubsection{Detailed TransPoint Schedule Designing}
\label{sec:method_transpoint}

\noindent
\textbf{TransPoint schedule from token length aspect.}\label{sec:method_transpoint_part1}
Due to the flexibility of TransPoint scheduling at different layers and token lengths, the model structure of TransMamba varies at different positions. Suppose the number of layers of the model is ${L}$ and the length of the sequence is ${T}$, there are ${L}*{T}$ possible TransPoint schedules for our TransMamba with fixed parameters.
We denote the value of TransPoint as ${P}$ (indicating that the tokens before position ${P}$ are modeled via Transformer and those after ${P}$ are encoded via Mamba), and we have the FLOPs for a TransMamba layer as follows:
\begin{equation}
\begin{aligned}
    \text{FLOPS}_{TransMamba} & = O({P}^2{N} + ({T} - {P}){N}^2).
\end{aligned}
\label{eq.flops_transmamba}
\end{equation}
Theoretically, the training time of TransMamba is a quadratic function of the TransPoint. 
In Section \ref{sec:exp:simple_transpoint_schedule}, our experiments confirm that the training efficiency of TransMamba indeed shows a quadratic function trend as TransPoint changes, and the optimal efficient point of our TransPoint $P$ is nearly $2,048$ for our setting ($N=1,536$ and $T=8,192$).

\noindent
\textbf{TransPoint schedule from layer aspect.}
We find that simply setting a global Transpoint for all layers (e.g., simultaneously at length 4,096) will result in unsatisfactory performance due to the sudden switching. To achieve better results, we need to set more diverse TransPoints at the layer level, which could gradually guide the model structure from pure Transformer to Mamba differently at various layers.
To enable a smoother transformation, the TransPoints are placed separately but as close as possible to the optimal efficient $P$ according to Eq. \ref{eq.flops_transmamba}.
Specifically, our TransPoints cycle is performed every 8 layers, referring to the work \citep{Mamba2} on hybrid structure. The mean of TransPoints is set slightly smaller than the optimal efficient point to enable more Mamba layers for better performance. TransPoints gradually transition from the beginning to the end of the sequence in a logarithmic trend (i.e., 0, 128, 256, 512, 1024, 2048, 4096, 8192), ensuring dispersion and smoothness.

Considering the above two aspects, we set our final TransPoint schedule balancing both effectiveness and efficiency. Our TransMamba with flexible TransPoint scheduling could have more potential interesting features to be further explored in the future. More detailed settings, explorations, and results are in the Experiments and Appendix.



\subsubsection{Diverse Inference Strategy}\label{sec:method:inf_diff}

Intuitively, the inference of TransMamba could adopt the same TransPoint schedule as that in training. However, due the flexibility of TransMamba, we can also choose completely different TransPoints during inference.
It provides us a whimsical but inspiring idea that we can train TransMamba with the most efficient structure, and choose a different structure that best suits the task during inference. In experiments, we will explore its potential.

\section{Experiments}

\subsection{Experiment Setup}\label{sec:exp:exp_setup}

We developed three baseline model families with various sizes (400M, 1.5B): Transformer \citep{shoeybi2019megatron}, Mamba2 \citep{Mamba2}, and Hybrid \citep{lieber2024jamba}.
All models are developed based on the Megatron-LM library.
Specifically, all models are unified in model size for fair comparisons.
The models are pre-trained utilizing collected in-house dataset which consists of a cleaned combination of Chinese and English datasets. We trained all models on 83 billion tokens for all models. For evaluation, we aim to achieve robust conclusions across diverse domains. We conducted comprehensive evaluations involving 8 English tasks, including ARC-E, ARC-C \citep{clark2018think}, CoQA \citep{reddy2019coqa}, OBQA \citep{mihaylov2018can}, PIQA \citep{bisk2020piqa}, PhoneBook \citep{waleffe2024empirical}, BoolQ \citep{clark2019boolq}, LongBench-v2 \citep{bai2024longbench}.
More details are shown in Appendix \ref{app:model_parameters_setting} and \ref{app:exp_setup_details}.


\subsection{Main Results}

\subsubsection{Evaluations on General Tasks}

\begin{table*}[!t]
  \centering
  \small
  \renewcommand\arraystretch{1.1}
  \setlength{\tabcolsep}{4.5pt}
  \begin{tabular}{lccccccc}
\toprule
    \textbf{Model}           & \textbf{ARC-E}& \textbf{ARC-C} & \textbf{CoQA}  & \textbf{OBQA} & \textbf{PIQA} & \textbf{PhoneBook}& \textbf{BoolQ} \\
    & \footnotesize\text{ACC} $\uparrow$& \footnotesize\text{ACC} $\uparrow$ & \footnotesize\text{F1-Score} $\uparrow$ & \footnotesize\text{ACC} $\uparrow$ & \footnotesize\text{ACC} $\uparrow$ & \footnotesize\text{Similarity} $\uparrow$& \footnotesize\text{ACC} $\uparrow$\\
\midrule
    Transformer-400M& 60.57 & \uline{58.72} & 5.07 & 42.4 & 52.75 & \uline{38.70}   &    60.72                 \\
    Mamba2-400M & 56.15 & 52.27 & 4.68  & 40.8 & 51.10 & 13.07 &       57.51               \\
    Hybrid-400M   & \uline{62.33} & 55.78  &  \uline{5.52} & \uline{43.6} & \uline{53.89}&  17.60        &       \uline{61.66}       \\
    \textbf{TransMamba-400M} & \textbf{62.50} & \textbf{59.33}& \textbf{6.23} & \textbf{44.8} & \textbf{55.76} &  \textbf{39.69}     &    \textbf{64.15}          \\
\midrule
    Transformer-1.5B & 60.87& \uline{59.43} & 5.93 & 48.6  & 56.66 & \textbf{41.04}& 61.42\\ 
    Mamba2-1.5B & 63.64 & 56.00 & 5.30 & 44.0 & 58.97 & 19.08 & 59.20\\ 
    Hybrid-1.5B  & \uline{63.92} & 57.97 & \uline{6.21} & \textbf{51.0} & \uline{59.25} & 26.63 & \uline{65.48} \\ 
    \textbf{TransMamba-1.5B}  & \textbf{64.75}& \textbf{63.33} & \textbf{6.97} & \uline{50.6}  & \textbf{59.61}  & \uline{40.92}& \textbf{66.73}\\
\bottomrule
  \end{tabular}
  \caption{Main evaluation results. TransMamba generally shows better performance.}
  \label{tab:main_eval}
\end{table*}

We evaluate the baselines and TransMamba on multiple tasks including question answering and reading comprehension as shown in Table \ref{tab:main_eval} (note that the input contexts of these tasks are longer enough than some of our TransPoints to trigger TransMamba). TransMamba achieves the overall best performance.
On the question answering and understanding tasks, TransMamba achieves the best performance or is comparable to the Hybrid model, while it consistently outperforms the original Transformer and Mamba2. 
The PhoneBook task is given the contact information of multiple people and requires the model to accurately answer the contact of a specific person. As introduced in Work \citep{waleffe2024empirical}, Mamba has a significant disadvantage compared to Transformer in this precise search task, and this disadvantage also brings to the Hybrid model. However, due to our smart combination of Transformer-Mamba at the sequence level, TransMamba can give accurate answers at the beginning of the sequence with almost the same accuracy as Transformer.

Table \ref{tab:exp:longbenchv2} shows the performance on the long-text benchmark LongBench-v2, where TransMamba still outperforms all baselines. This further illustrates the role of the lossless Memory Converter in TransMamba, which can effectively preserve the information before TransPoint.

\begin{table}[!t]
\small
  \begin{minipage}{0.44\textwidth}
    \begin{center}
  \renewcommand\arraystretch{1.1}
  \setlength{\tabcolsep}{4.3pt}
\begin{tabular}{lccc}
\toprule
\multirow{2}{*}{\textbf{Model}}          & \multicolumn{3}{c}{LongBench-v2}
\\ & Overall & Easy & Hard 
\\
\midrule
Transformer          &   31.61   &  34.38   & 29.90      \\
Mamba                &   30.62   &  32.81   &  29.26     \\
Hybrid               &    35.79  &  38.02   &  34.41     \\
\textbf{TransMamba}  &    \textbf{38.76}  &  \textbf{40.10}   &   \textbf{37.94}    \\
\bottomrule
\end{tabular}
\end{center}
\captionof{table}{Evaluation results of our TransMamba and baselines on the long text benchmark LongBench-v2. The number of parameters of all models is 1.5B.}\label{tab:exp:longbenchv2}
  \end{minipage}
  \hfill
  \begin{minipage}{0.5\textwidth}
    \begin{center}
  \renewcommand\arraystretch{1.1}
  \setlength{\tabcolsep}{3.5pt}
\begin{tabular}{lcc}
\toprule
\multirow{2}{*}{\textbf{Model}}          & \textbf{Relative}  & \textbf{Flops} /  \\ & \textbf{Train Time} & \,\, (Layer,  \num{e10})
\\
\midrule
Transformer &   1.00   &   10.51    \\
Mamba &   0.77    &   2.01  \\
Hybrid &  0.78     &   6.26 \\
\textbf{TransMamba} &  \textbf{0.75}    &   \textbf{1.91}   \\
\bottomrule
\end{tabular}
\end{center}
\captionof{table}{Comparison of average training time of baseline and TransMamba. Relative time refers to the ratio of the time to train the same batch-size of the baseline to Transformer.}\label{tab:main_efficiency_exp}
  \end{minipage}%
\end{table}

\subsubsection{Efficiency Analysis}



As described in Table \ref{tab:efficiency_attn_SSM} and Section \ref{sec:effiency_of_attn_SSM}, Transformer and Mamba have efficiency advantages on short and long text, respectively. 
Our TransMamba is more efficient compared to baselines.
Specifically, taking sequence length T=8k and state dimension N=4k as an example, the theoretical FLOPs of Transformer is 2.29 times that of the optimal TransMamba, while that of Mamba is 1.14 times.
We conducted experiments on the average training time of the baselines and TransMamba on 3 machines in Table \ref{tab:main_efficiency_exp}. TransMamba has a maximum efficiency improvement of 25\% compared to Transformer, which will increase to 0.8\% if we utilize the optimal efficient TransPoint Schedule. This efficiency improvement is consistent with the relative size of the theoretical FLOPs value.
It is worth noting that because attention and SSM have their own engineering acceleration, the actual runtime improvement result does not fully reach the theoretical speedup limit. Optimization of TransMamba acceleration engineering is our future work, and there is still potential for speed improvement.

\subsection{In-depth Analyses on Different TransPoint Schedule}

\subsubsection{Analyses on Layer-Shared TransPoint Schedule}\label{sec:exp:simple_transpoint_schedule}

\begin{table}[!t]
\centering
\small
  \renewcommand\arraystretch{1.3}
  \setlength{\tabcolsep}{5.5pt}
\begin{tabular}{cc|c|cc}
\hline
 \multicolumn{2}{c|}{\multirow{2}{*}{Model Setting}} & \multirow{2}{*}{Detailed TransPoint Schedule} & \multicolumn{2}{c}{Validation} \\
                                   & &                                               & Loss $\downarrow$           & PPL $\downarrow$          \\ \hline
  \multicolumn{2}{c|}{Transformer} & [8192]   & 3.098 & 2.194                \\ \hline
 \multirow{3}{*}{Layer-shared} & V1 & [2048] & 3.356 & 2.401\\
 & V2 & [4096]& 3.297& 2.346 \\
 & V3 & [6144]& 3.308 & 2.339 \\ \hline
         \multirow{3}{*}{Layer-specific} &  V4           & [3072, 4096, 5120]                            & 3.125          & 2.287         \\
                                      &  V5        & [2048, 3072, 4096]                            & 3.100          & 2.219         \\
                                       & V6        & [512, 1024, 2048]                             & 3.135          & 2.299         \\ \hline
     \multirow{2}{*}{Broad-range} &    V7           & [2048, 4096, 6144]                            & 3.084          & 2.185         \\
                                   &      V8            & [0, 1024, 2048, 6144, 8192]                   & 3.022          & 2.053         \\ \hline
Fine-grained  & V9                           & [0, 128, 256, 512, 1024, 2048, 4096, 8192]      & \textbf{2.898}          & \textbf{1.813}         \\ \hline
\end{tabular}
\caption{Results of different TransPoint schedule. The input token sequence length of the training data is 8192. The validation loss and PPL is calculated at 21 billion tokens. The TransPoint of each layer in the model cyclically alternates through the predefined TransPoints sequence with the pattern repeating.}
\label{tab:exp:complex_transpoints}
\end{table}





TransPoint scheduling has a significant impact on the effectiveness and efficiency. We first conducted experiments on Layer-shared TransPoint scheduling for a straightforward understanding, where the TransPoint of all layers is set to one unified value. We experimented with the training efficiency of the TransPoint setting from 0 to 8192 in step size of 64.
As shown in Figure \ref{fig:exp_simple_transpoints} (a), the relative training time shows a quadratic curve trend, and the optimal TransPoint is around $2,048$.
V1 $\sim$ V3 in Table \ref{tab:exp:complex_transpoints} shows different Layer-shared schedules at various positions, whose loss and PPL are not satisfactory due to the sudden mutation from Transformer to Mamba for all layers. Hence, we move to explore Layer-specific TransPoint scheduling for better performance.

\subsubsection{Analyses on Layer-specific TransPoint Schedule}\label{sec:exp:complex_transpoint_schedule}

\renewcommand{\thefigure}{\arabic{figure}} 
\renewcommand{\thesubfigure}{(\alph{subfigure})} 
\begin{figure}[t]
    \centering
    \begin{minipage}[t]{0.46\textwidth}
        \centering
        \includegraphics[width=\textwidth]{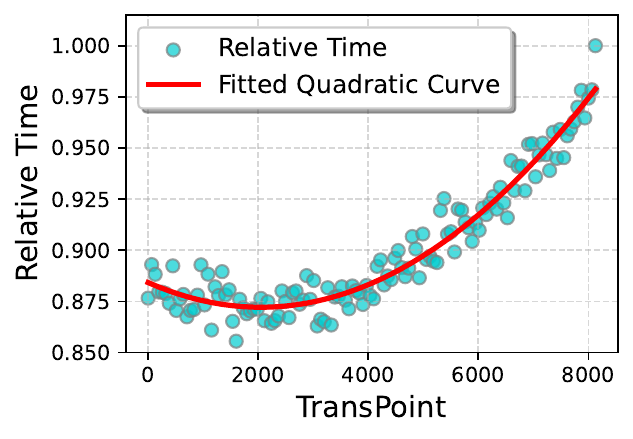}
        \captionof{subfigure}{Layer-Shared TransPoint Schedule}
        \label{fig:img1}
    \end{minipage}
    \hfill
    \begin{minipage}[t]{0.46\textwidth}
        \centering
        \includegraphics[width=\textwidth]{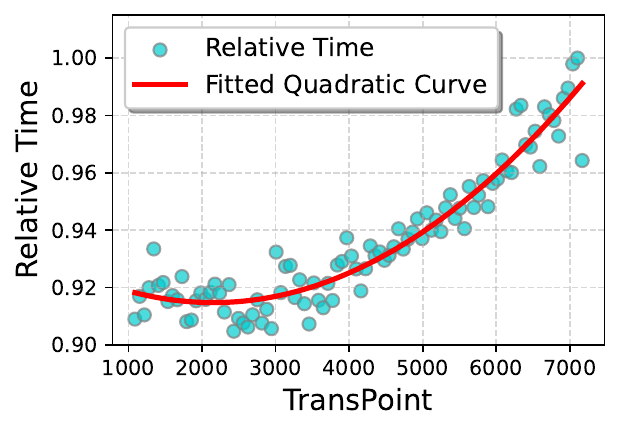}
        \captionof{subfigure}{Layer-Specific TransPoint Schedule}
        \label{fig:img2}
    \end{minipage}
    \caption{Experiments on TransPoint Schedule and training efficiency. }
    \label{fig:exp_simple_transpoints}
\end{figure}

During explorations on Layer-specific scheduling, we observed that three characteristics can bring better results: \textbf{\emph{Layer-specific}}, \textbf{\emph{Broad-range}}, and \textbf{\emph{Fine-grained}}, and conduct three groups of evaluations (V4 $\sim$ V9 in Table \ref{tab:exp:complex_transpoints}). For example, the schedule of V4 indicates that its TransPoints cycle every 3 layers, and the TransPoints of layer 1-6 are: [3072, 4096, 5120, 3072, 4096, 5120...]. Note that these Layer-specific schedules also possess the same quadratic curve trend on efficiency as shown in Figure \ref{fig:exp_simple_transpoints} (b). We condensed the following rules for TransPoints scheduling, and the final schedule is based on both effectiveness and efficiency.
\noindent
\textbf{(a) The TransPoints of each layer should be \emph{layer-specific}.} Setting concentrated TransPoints for all layers has relatively poor validation results. V4 $\sim$ V6 set TransPoints at three positions and achieve better loss and PPL compared to V1 $\sim$ V3 with shared TransPoints.
\noindent
\textbf{(b) The scheduling of TransPoints should cover \emph{broad-range} of the sequence.} V4 $\sim$ V6 have TransPoints of all layers under the concentrated setting vary within the range of 2k tokens, while the verification loss and PPL are significantly higher compared to V7 and V8. More diverse TransPoint help the model performance gradually improve. 
\noindent
\textbf{(c) \emph{Fine-grained} transformation of TransPoints improves the performance.} Compared to the vanilla broad range setting V7 and V8, V9 (i.e., the final TransMamba setting) have finer-grained and smoother scheduling cycling every 8 layers, achieving the best result.
\subsection{Explorations on Inconsistent Training/Inference TransPoint Scheduling}\label{sec:exp:inf_diff}

Due to the flexibility of the Transformer and Mamba mode transformation based on the unified parameters in TransMamba, we can set completely different TransPoint schedules for training and inference. 
For this bold exploration, we train our TransMamba with the selected schedule V9, and then inference with different schedules. 
 We surprisingly find that some inconsistent training/inference settings (e.g., inference with Transformer) could not only function normally, but also achieve even better results on certain tasks. We present these charming results in the Appendix \ref{app:inf_diff}, which is a promising research direction.



%

\subsection{Ablation Study on Other Model Components}
We conduct experiments on the components of the details of TransMamba framework. We found it beneficial to radiate the key components of Mamba onto the overall structure of TransMamba. The experimental results and details are shown in Appendix \ref{app:Model_Components}.

\section{Related Works}
Transformer has always been the focus of language model research \citep{beltagy2020longformer, liu2021swin, tang2024survey}, but its limitations in processing long sequences \citep{zhou2021informer, behrouz2024titans} and the memory pressure caused by KV cache \citep{wang2020linformer, dao2022flashattention} are also difficult to solve. Mamba has the advantage of linear complexity based on the state space model \citep{mamba, zhang2024survey}, but struggles in modeling complex contexts \citep{xiao2024spatial}.

Hybrid models \citep{chen2024dsdformer, lou2024sparx, ren2025vamba} that combine the two are emerging, but most of the work simply cascades them \citep{hatamizadeh2024mambavision, lieber2024jamba}. Recent work \citep{Mamba2, han2024demystify, distill_mamba} has revealed the consistency of the underlying mathematics between them. However, there is no work that truly attempts to unify Transformer and Mamba in sequence level.

\section{Conclusion}
We proposes TransMamba to unify Transformer and Mamba at the sequence level and proves its superiority in efficiency and performance. Furthermore, we conduct a detailed exploration of TransPoints and summarize three criteria of TransPoint Scheduling. 
In short, our attempt provides insight and inspiration for the next generation of sequence modeling.




\bibliography{colm2025_conference}
\bibliographystyle{colm2025_conference}

\newpage

\appendix
\section{Appendix}
\subsection{Detailed Conclusion, Future Work and Limitations}\label{app:future_work}

This paper proposes TransMamba to unify Transformer and Mamba at the layer level and proves its superiority in efficiency and performance. Specifically, we combine the advantages of Transformer and Mamba in long and short contexts to significantly improve the training speed during training. At the same time, conversion modules such as Memory Converter ensure lossless model conversion and ensure the performance of the model.

We conduct a detailed exploration of TransPoints and model framework, from naive layer-shared TransPoint scheduling to sophisticated layer-specific design, and summarize three standards of the TransPoint  scheduling. At the same time, due to the flexible model architecture of TransMamba under shared parameters, we boldly tried training and reasoning isomorphism and obtained surprising results. In short, our attempt provides insight and inspiration for the next generation of sequence modeling.

This paper also has some limitations for future research, including:

\begin{enumerate}
    \item Future work could try larger models and explore the form of the scaling law of TransMamba;
    \item Since Transformer and Mamba have different degrees of optimization, the actual optimal value of TransPoint has the potential to be further explored. In our current experiments, we concluded that the degree of training optimization of Transformer and Mamba is proportional, and the proportionality coefficient is approximately Transformer: Mamba=2.67:1 (i.e., the training speed of Transformer will be 2.67 times faster than that of Mamba under the same FLOPs). This provides ideas for follow-up work;
    \item Transformer and Mamba have their own variants. Follow-up work can try to combine different variants into a new TransMamba. This research direction has a lot of possible exciting results.
\end{enumerate}

\subsection{Method Details}
\subsubsection{Memory Converter}\label{app:method_details_MemoryConverter_1}
The matrix utilized in Equation ~\ref{eq:used_A} is as follows:
\begin{equation}
 \mathbf{A}^{\times} = \begin{bmatrix}
1 &  &  &  \\
\overline{\mathbf{A}}_1 & 1 &  &  \\
\overline{\mathbf{A}}_2\overline{\mathbf{A}}_1 & \overline{\mathbf{A}}_2 & 1 &  \\
\overline{\mathbf{A}}_3\overline{\mathbf{A}}_2\overline{\mathbf{A}}_1 & \overline{\mathbf{A}}_3\overline{\mathbf{A}}_2 & \overline{\mathbf{A}}_3 & 1 
\end{bmatrix}. 
\end{equation}
\subsection{Experiment Details}

\subsubsection{Model Parameters Setting}\label{app:model_parameters_setting}

\begin{table}[h]
\centering
  \renewcommand\arraystretch{1.1}
\begin{tabular}{cc}
\hline
Settings          & Value   \\ \hline
Global Batch Size & 1024    \\
Micro Batch Size  & 2       \\
Sequence Length   & 8192    \\
Train Tokens      & 81B     \\
MLP Ratio         & 0.5     \\
Initial lr        & 2.5e-4  \\
Min lr            & 2.5e-5  \\
lr-Decay-Style    & cosine  \\
weight-decay      & 0.1     \\
clip-grad         & 1.0     \\
Normalization     & RMSNorm \\
adam-beta1        & 0.9     \\
adam-beta2        & 0.95    \\
bf16              & True    \\
rope-theta        & 10000   \\ \hline
\end{tabular}
\caption{Global parameter settings.}
\label{tab:app:global_model_settings}
\end{table}

\begin{table}[!h]
\centering
  \renewcommand\arraystretch{1.2}
\begin{tabular}{cc|c}
\hline
\multicolumn{2}{c|}{Model Setting}          & Value \\ \hline
\multirow{4}{*}{400M} & Num-Layers          & 24    \\
                      & Hidden-Size         & 1536  \\
                      & FFN-Hidden-Size     & 4096  \\
                      & Num-Attention-Heads & 16    \\ \hline
\multirow{4}{*}{1.5B} & Num-Layers          & 64    \\
                      & Hidden-Size         & 1536  \\
                      & FFN-Hidden-Size     & 4096  \\
                      & Num-Attention-Heads & 16    \\ \hline
\end{tabular}
\caption{Model parameter setting.}
\label{tab:app:model-setting}
\end{table}

In this section we introduce the parameter settings of the baseline and TransMamba used in this paper. Empirically, we set the ratio of TransMamba layer to MLP layer to 1:1. (The baseline also has the same setting). Table \ref{tab:app:global_model_settings} shows the overall training settings, and Table \ref{tab:app:model-setting} shows the specific parameters of each model size.

\subsubsection{Experiment Setup Details}\label{app:exp_setup_details}

The benchmarks used in this paper introduced in Section \ref{sec:exp:exp_setup} are described as follows:
\begin{itemize}[leftmargin=0.36cm]
    \item ARC \citep{clark2018think} dataset, developed by the Allen Institute for Artificial Intelligence (AI2), is a collection of 5,197 elementary-level science questions designed to evaluate natural language understanding and reasoning capabilities in AI systems, focusing on straightforward scientific concepts typically encountered in grade school curricula.
    \item CoQA \citep{reddy2019coqa} dataset is a large-scale collection of 127,000 question-answer pairs from 8,000 dialogues across seven diverse domains (e.g., news, literature, science), designed to evaluate machines' ability to answer context-dependent, free-form questions in multi-turn conversations while requiring coreference resolution and pragmatic reasoning.
    \item OBQA \citep{mihaylov2018can} dataset is a novel question-answering benchmark designed to evaluate AI systems' ability to integrate external commonsense knowledge and perform multi-step reasoning, requiring comprehension beyond direct text retrieval to answer science-based questions aligned with elementary school curricula.
    \item PIQA \citep{bisk2020piqa} dataset is a benchmark designed to evaluate AI systems' reasoning capabilities about physical commonsense knowledge through context-dependent questions that require understanding object properties, manipulation strategies, and real-world physics (e.g., "How to separate egg yolk using a water bottle?"), with human accuracy reaching 95\% while state-of-the-art models achieve ~77\% accuracy.
    \item PhoneBook is introduced in \citep{waleffe2024empirical} and aims to evaluate the exact phone number of a specific person given a phone book of multiple people. There are two ways to construct a specific phone book: conventional construction and reverse construction. We use the open source tool \citep{Faraglia_Faker} to construct a completely random test set PhoneBook.
    \item  BoolQ \citep{clark2019boolq} dataset is a natural language understanding benchmark comprising 15,942 yes/no questions paired with contextual paragraphs, designed to evaluate models' ability to answer binary questions through complex reasoning over real-world web content, where human performance reaches 90\% accuracy while models initially struggled to surpass 70\%.
    \item LongBench-v2 \citep{bai2024longbench} dataset is a comprehensive multilingual benchmark designed to evaluate large language models' (LLMs) deep understanding and reasoning capabilities in ultra-long contexts (8k–2M words) through 503 challenging multiple-choice questions spanning six task categories, including single/multi-document QA, long in-context learning, and code repository analysis, with human expert accuracy limited to 53.7\% under constrained conditions.
\end{itemize}

\subsection{Additional Experiments of Ablation Study}
\subsubsection{Explorations on Inconsistent Training/Inference TransPoint Scheduling}\label{app:inf_diff}

In this section, we supplement the experimental results of inconsistent training/inference TransPoint scheduling. Note that this setting is extreme challenging. As shown in Table \ref{tab:app:Inconsistent Inference TransPoint}, TransMamba can still maintain a certain level in most cases under completely different reasoning structures. Even in some cases, such as the score of OBQA evaluated with the Hybrid structure, it can exceed the original TransMamba and all baselines. This gives us a lot of inspiration for future research directions. The structural decoupling of reasoning can bring many research possibilities and unexplored performance.

\begin{table*}[h]
  \centering
  \small
  \renewcommand\arraystretch{1.1}
  \begin{tabular}{lccc}
\toprule
    \textbf{Model}           & \textbf{ARC-E}  & \textbf{OBQA} & \textbf{PIQA}  \\
    & \footnotesize\text{ACC} $\uparrow$  & \footnotesize\text{ACC} $\uparrow$ & \footnotesize\text{ACC} $\uparrow$ \\
\midrule
    Transformer-400M& 60.57  & 42.4 & 52.75   \\
    Mamba2-400M & 56.15     & 40.8 & 51.10                       \\
    Hybrid-400M   & \uline{62.33}    & \uline{43.6} & \uline{53.89}                      \\
    \textbf{TransMamba-400M} & \textbf{62.50}  & \textbf{44.8} & \textbf{55.76}                    \\
\midrule
    TransMamba-400M-Inf @ Transformer & 27.27 & 34.8 &  50.49   \\ 
    TransMamba-400M-Inf @ Mamba2      & 50.82  & 38.2  & 50.89   \\ 
    TransMamba-400M-Inf @ Hybrid      & 52.20 & \textbf{45.0} &  51.30   \\ 
\bottomrule
  \end{tabular}
  \caption{Results of inconsistent training/inference TransPoint scheduling. Although the ``Inf'' TransMamba versions perform worse than the original consistent version in bold, the close performance inspires us to conduct future explorations.}
  \label{tab:app:Inconsistent Inference TransPoint}
\end{table*}

\subsubsection{Model Components}\label{app:Model_Components}


In the process of building TransMamba, we conducted detailed experiments on each component of the model. Table \ref{tab:app:model_comonents} shows some of the key results. Our most critical conclusions include: (1) In TransMamba, the attention block is not suitable for mapping with the z of SSM; (2) Memory Converter optimization is necessary. After we modified from a simple MLP fitting to the theoretical solution Memory Converter, the running speed and training effect of the model were significantly improved. In addition, the SSM block acceleration mentioned in the Mamba paper can also be used for Memory Converter.

\begin{table}[h]
\centering
  \renewcommand\arraystretch{1.1}
\begin{tabular}{cc|c}
\hline
\multicolumn{2}{c|}{Experiment}                                                                                   & Training Loss \\ \hline
\multicolumn{1}{c}{\multirow{2}{*}{Global z}} & h                                                                & 3.503      \\
\multicolumn{1}{c}{}                          & residual                                                         & 3.447      \\ \hline
\multicolumn{2}{c|}{Attention w/o z}                                                                              & 3.39       \\ \hline
\multirow{2}{*}{Memory Converter}              & MLP                                                              & 3.209      \\
                                               & \begin{tabular}[c]{@{}c@{}}SSM \\ (Current Version)\end{tabular} & 3.173      \\ \hline
\end{tabular}
\caption{The training losses of ablation versions with different model structures.}
\label{tab:app:model_comonents}
\end{table}


\end{document}